\documentclass{article}

    \PassOptionsToPackage{numbers, compress}{natbib}
    \usepackage[preprint]{neurips_2019}

\usepackage[utf8]{inputenc} % allow utf-8 input
\usepackage[T1]{fontenc}    % use 8-bit T1 fonts
\usepackage{hyperref}       % hyperlinks
\usepackage{url}            % simple URL typesetting
\usepackage{booktabs}       % professional-quality tables
\usepackage{amsfonts}       % blackboard math symbols
\usepackage{nicefrac}       % compact symbols for 1/2, etc.
\usepackage{microtype}      % microtypography
\usepackage{graphicx}
\usepackage{subcaption}
\usepackage{amsmath}

\title{Hyperparameter Analysis for Image Captioning}

\author{%
  Amish Patel \\
  Department of Computer Science\\
  University of Toronto\\
  \texttt{amishpatel@cs.toronto.edu} \\
   \And
   Aravind Varier \\
   Department of Computer Science\\
   University of Toronto\\
  \texttt{avarier@cs.toronto.edu} \\
}

\begin{document}

\maketitle

\begin{abstract}
  In this paper, we perform a thorough sensitivity analysis on state-of-the-art image captioning approaches using two different architectures: CNN+LSTM and CNN+Transformer. Experiments were carried out using the Flickr8k dataset. The biggest takeaway from the experiments is that fine-tuning the CNN encoder outperforms the baseline and all other experiments carried out for both architectures.
\end{abstract}

\section{Introduction}
Image captioning is a popular research area as it combines the domains of Computer Vision and Natural Language Processing. Generating captions automatically is a difficult problem since it not only involves detecting the objects present in the image, but also involves expressing the semantic relationship between the corresponding objects in a natural language. 

As the deep learning community matures, there have been several approaches for image captioning which have produced state-of-the-art models that produce sentences closer to natural language \cite{show-attend-tell}\cite{show-and-tell}. The different approaches usually follow the general encoder-decoder structure. The role of an encoder is to extract the semantic information from the image and is typically represented by Convolutional Neural Networks (CNN). The role of a decoder is to translate the encoded image features into natural language and this is usually represented by Recurrent Neural Network (RNN) \cite{show-attend-tell}. In particular, a special type of RNN is popular and is referred to as Long Short Term Memory (LSTM) which has the ability to handle long-term temporal dependencies better through the use of a memory cell. However, RNN is known to be sequential in the sense that it generates one word at a time and as such, the training time increases with increased sequence length of the caption. As such, the transformer architecture has been used in recent times in place of an RNN \cite{transformer-stacked-attention} for image captioning to avoid this sequential training problem. 

Our key contribution in this paper is that we perform sensitivity analysis of several hyperparameters for both LSTM-based decoder and Transformer-based decoders by using Flickr8k dataset.

\subsection{Related Work}
After achieving a lot of success in Neural Machine Translation tasks, the encoder-decoder framework has been used several times for image captioning \cite{show-attend-tell}\cite{transformer-stacked-attention}\cite{show-and-tell}. \citet{show-and-tell} used CNN as the encoder for the image and then used Long Short Term Memory (LSTM) RNN as the decoder. In particular, they only used the encoded image representation for the first timestep for the LSTM. \citet{show-attend-tell} extended this approach by proposing a visual attention mechanism to attend to different parts of the images for every timestep during the caption generation using LSTM. 
% In particular, the attention mechanism they used was introduced by \citet{bahdanau-attention-nmt} and the base idea is that the attention weights are computed via a multi-layer perceptron (MLP) conditioned on previous hidden state and encoded representation of the image. This is worth mentioning as our RNN-based decoder architecture for the sensitivity analysis is directly based on the Soft-Attention model introduced by \citet{show-attend-tell}. 
Beyond the recurrence-based decoders, \citet{transformer} established a new architecture for machine translation called Transformer which is purely based on attention mechanism. They also showed that a Transformer is superior is quality while taking significantly less time to train as it is parallelizable. Using Transformer for image captioning has achieved state-of-the-art results as shown by \citet{transformer-stacked-attention}.

\section{Approach}
In this section, we discuss the methodology used for our experiments. In particular, we used an encoder-decoder architecture for image captioning where the encoder consisted of a CNN and two different decoder networks were examined: LSTM and Transformer.

\subsection{Flickr8k Dataset}
The dataset used for experimentation is Flickr8K \cite{image-captioning-survey} which has 8,000 images in total. In particular, it is divided in to 6,000 training images, 1,000 validation images, and 1,000 test images. Furthermore, each of the images is associated with five reference captions annotated by humans. As such, our training set consists of 30,000 samples where each sample corresponds to one image and one caption. 

\subsection{Encoder}
The encoder model we used was the ResNet CNN model. The CNN extracts the features of the image which are referred to as annotation vectors. These vectors form the hidden states of the encoder on which the attention mechanism is performed. We experimented with 3 different types of ResNet models, the ResNet18, ResNet50 and the ResNet101. The number following the model name indicates the number of layers. We removed the final pooling and softmax layer and extracted the features from the final convolutional layer. We obtain an output of size N x 14 x 14, where the value of N depends on the type of encoder used. This is then flattened to give us a 196-dimensional vector on which we perform attention.

\subsection{LSTM Decoder}
The LSTM network produces a caption by generating a word at every time-step. The output at a given time-step is conditioned on the current hidden state, a context vector which is obtained from the attention mechanism and all the previous hidden states. The initial hidden state and cell state of the LSTM is obtained by taking the average of the annotation vectors and passing it through different MLPs. At a given time-step, we perform attention to obtain a context vector which is then appended to the input word embedding. In particular, we follow the same soft attention training process that is described by \citet{show-attend-tell} for our experiments.

% The context vector is obtained according to the following formula:

% %formula

% Following this, the output word at the current time step is calculated using the deep output layer. This is given as:

% %formula

% During training, the correct word is fed in as input at every timestep. This is known as teacher forcing. At the end of the sample, we calculate the total loss by considering two individual losses. The first is the cross entropy loss. At each timestep the model predicts a word with a certain probability. This is used to calculate the cross entropy loss. Along with this, a second loss is calculated called the Doubly Stochastic loss. The way the attention weights are calculated %formula
% , it adds up to 1 due to the softmax. This additional loss ensures that %formula
% This ensures that the model pays equal attention to all regions of the image during training.

\subsubsection{Transformer Decoder}

\begin{figure}[htpb!]
  \centering
  \includegraphics[width=0.4\linewidth]{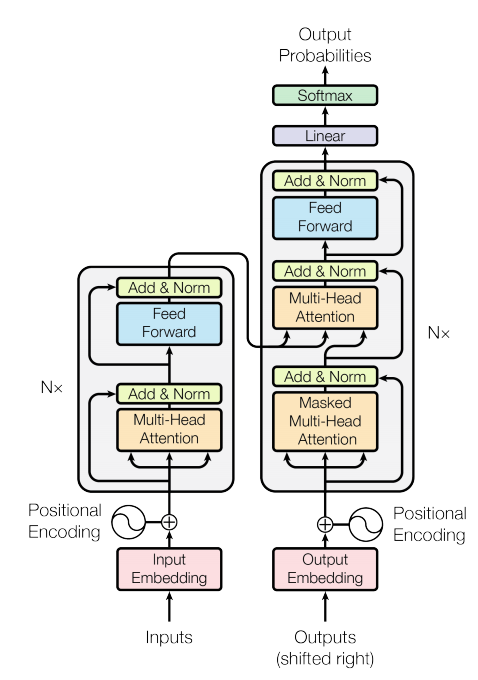}
  \caption{Transformer architecture. We substitute the encoder with a CNN.}
  \label{fig:transformer}
\end{figure}

The transformer model introduced by \citet{transformer} was a way forward for language modeling that did away with the recurrent nature of language modeling. It relied purely on self-attention. The transformer model is shown in Figure \ref{fig:transformer}.

\begin{figure}[htpb!]
    \centering
    \begin{subfigure}[b]{0.4\linewidth}
        \includegraphics[width=\linewidth]{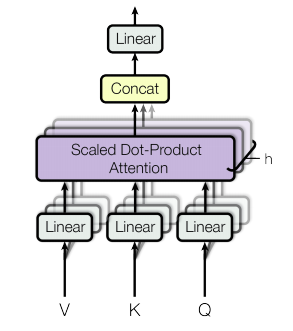}
        \caption{Multi-head attention.}
        \label{subfig:multihead}
    \end{subfigure}
    \begin{subfigure}[b]{0.4\linewidth}
        \centering
        \includegraphics[width=0.5\linewidth]{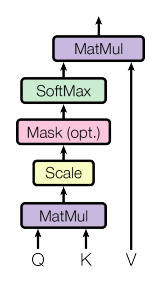}
        \caption{Scaled dot-product attention function.}
        \label{subfig:attention}
    \end{subfigure}
    \caption{Attention mechanisms used in a Transformer.}
    \label{fig:multihead_&_attention}
\end{figure}

We use the encoder as the CNN and adopt just the decoder of the transformer architecture. The transformer network relies on a series of computations known as scaled dot-product attention. The attention function is basically a mapping of queries(Q) and key-value pairs(K-V) to an output. The output is a weighted sum of the values, where the weights assigned to each value is based on a similarity function between the query and key. This is shown in Figure \ref{subfig:attention}. It is given by the following formula:
\begin{equation}
    Attention(Q, K, V) = softmax(\frac{QK^T}{\sqrt{d_k}})V
\end{equation}
Here $\sqrt{d_k}$ is a scaling factor that prevents the absolute value of the dot product from blowing up.

The transformer employs another strategy known as Multi-Head self-attention. This ensures that the model learns a multi-modal representation of the input sentence. The model learns to attend to different representations of the same input. The idea is to project the input vectors to different sub-spaces followed by the self attention function in each subspace. The output of each subspace is concatenated and a linear layer projects the data back down to the original subspace. This is shown in Figure \ref{subfig:multihead}. The formula is given by:
\begin{align*}
    MultiHead(Q, K, V) &= Concat(head_1, ..., head_h)W^O \\ where, head_i &= Attention(QW_i^Q, KW_i^K, VW_i^V)\\
\end{align*}
The decoder block can be broken into three main sub-blocks. The first is a masked multi-head self attention layer. This is a layer of self attention where the only difference is that each vector only attends to words that come before it. This is so that the model only uses past information to make judgements about the present and does not get a peek into the future. The second sub-layer is a layer of multi-head attention on the encoder hidden states. This is where the representation of the image is fed into the decoder layer. Finally, there is a feed forward layer which introduces some non-linearity to the model.
% \begin{itemize}
%     \item The first is a masked multi-head self attention layer. This is a layer of self attention where the only difference is that each vector only attends to words that come before it. This is so that the model only uses past information to make judgements about the present and does not get a peek into the future. 
%     \item The second sub-layer is a layer of multi-head attention on the encoder hidden states. This is where the representation of the image is fed into the decoder layer. 
%     \item Finally there is a feed forward layer which introduces some non-linearity to the model.
% \end{itemize}

In between each sub-layer there is a residual connection which speeds up convergence and prevents the vanishing gradient problem. There are also dropout layers after each sub-layer to prevent over-fitting. Multiple such decoder blocks are stacked one on top of the other until the outputs of the last layer are passed through a softmax layer to obtain the output probabilities. Another technique for regularization we used was label-smoothing with $\epsilon=0.1$. It was noticed that the model had higher loss but the BLEU scores improved.

\subsection{Metrics}
We use BLEU \cite{bleu}, METEOR \cite{meteor}, CIDEr \cite{cider}, ROUGE \cite{rouge} to evaluate the quality of generated captions. BLEU \cite{bleu} measures similarity between a set of reference texts and the machine generated text through the use of n-grams. METEOR \cite{meteor} is based on explicit word to word matches through the use of corresponding word stems and synonyms. CIDEr \cite{cider} uses Term Frequency-Inverse Document Frequency (TF-IDF) weighting for each n-gram to measure similarity between reference texts and predicted text. ROUGE \cite{rouge} uses word pairs, n-grams, and word sequences to measure sentence similarity. Existing image captioning research use BLEU, METEOR, and ROUGE extensively. However, CIDEr has been found to be more correlated with human assessment \cite{image-captioning-survey}. As such, we decided to include CIDEr as well to get a better representation of caption quality.

\section{Experimental Analysis}
We performed several experiments to analyze the difference in using different encoder-decoder architectures.
\subsection{ResNet + LSTM}
Our baseline model was ResNet18 combined with an LSTM using a hidden vector size of 512. We do not finetune the encoder in the baseline. The word embedding size used in all LSTM experiments are of size 512. We conducted three different experiments: varying the encoder, fine-tuning the encoder, and varying the number of LSTM hidden units. All experiments were trained using the Adam optimizer with a learning rate of 0.0001 on Nvidia GeForce GTX 1080 GPU. In additon, the termination of training was determined by early stopping to obtain the best possible BLEU-4 scores.

\begin{figure}[htpb!]
  \centering
  \includegraphics[width=0.5\linewidth]{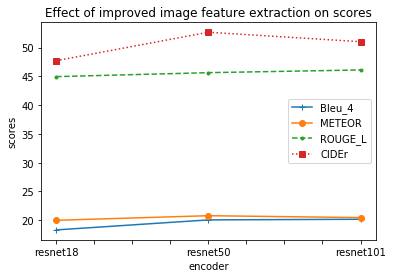}
  \caption{Effect of varying encoders on image captioning for CNN+LSTM architecture.}
  \label{fig:exp-lstm-image-quality}
\end{figure}

For the first experiment, we used ResNet18, ResNet50, and ResNet101 models in order to examine the effect of improved image quality. It was hypothesized that using a larger CNN model would result in better caption generation. The summary of the experimental results is listed in Table \ref{tab:lstm-varying-encoder}. Moreover, Figure \ref{fig:exp-lstm-image-quality} shows that ResNet50 and ResNet101 perform much better than ResNet18 and hence, validates our original hypothesis. One key observation is that performance difference between RestNet50 and ResNet101 is minimal and ResNet101 actually gave a lower CIDEr score. As such, there is an indication that increasing image quality after some point may lead to saturation in terms of caption quality.

\begin{table}[htpb!]
  \caption{Experimental results for varying encoder in CNN+LSTM architecture.}
  \label{tab:lstm-varying-encoder}
  \centering
  \begin{tabular}{llllllll}
    \toprule
    Encoder & BLEU-1 & BLEU-2 & BLEU-3 & BLEU-4 & METEOR & ROUGE\_L & CIDEr \\
    \midrule
    ResNet18 & 58.57 & 40.37 & 27.39 & 18.32 & 19.99 & 44.94 & 47.71 \\
ResNet50 & 59.95 & 42.34 & 29.26 & 20.06 & 20.79 & 45.63 & 52.65 \\
ResNet101 & 60.73 & 43.06 & 29.55 & 20.17 & 20.46 & 46.10 & 51.01 \\
    \bottomrule
  \end{tabular}
\end{table}

\begin{figure}[htpb!]
  \centering
  \includegraphics[width=0.5\linewidth]{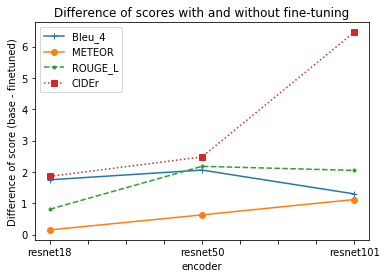}
  \caption{Effect of fine-tuning on image captioning for CNN+LSTM architecture. Positive difference indicates that fine-tuned model outperforms the base model without fine-tuning.}
  \label{fig:exp-lstm-finetuned}
\end{figure}

Next, we decided to fine-tune all three encoders discussed above and see it's effect on caption quality. It was hypothesized that fine-tuning would increase captioning quality as ResNet is trained on ImageNet while we are using Flickr8k. For each of the encoders, we compared the score of fine-tuned model with the corresponding model scores without fine-tuning. The summary of the results is listed in Table \ref{tab:lstm-finetuned}. Moreover, Figure \ref{fig:exp-lstm-finetuned} shows that fine-tuning is beneficial for all encoders as it outperforms the base models. One important observation is that the deeper models benefit more from fine-tuning as it can be clearly seen from the positive trend for CIDEr and METEOR scores. Additionally, we were able to outperform the metrics reported by \citet{show-attend-tell} through these experiments.  

\begin{table}[htpb!]
  \caption{Experimental results for fine-tuning last CNN layer in CNN+LSTM architecture. It shows the difference of scores by subtracting the base model scores from fine-tuned scores.}
  \label{tab:lstm-finetuned}
  \centering
  \begin{tabular}{llllllll}
    \toprule
    Encoder & $\Delta$ BLEU-1 & $\Delta$ BLEU-2 & $\Delta$ BLEU-3 & $\Delta$ BLEU-4 & $\Delta$ METEOR & $\Delta$ ROUGE\_L & $\Delta$ CIDEr \\
    \midrule
    ResNet18 & 1.95 & 2.00 & 1.84 & 1.75 & 0.15 & 0.81 & 1.86 \\
ResNet50 & 2.74 & 2.75 & 2.60 & 2.06 & 0.63 & 2.18 & 2.48 \\
ResNet101 & 2.14 & 2.07 & 1.92 & 1.30 & 1.12 & 2.05 & 6.46 \\
    \bottomrule
  \end{tabular}
\end{table}

\begin{figure}[htpb!]
  \centering
  \begin{subfigure}[b]{0.45\linewidth}
    \includegraphics[width=\linewidth]{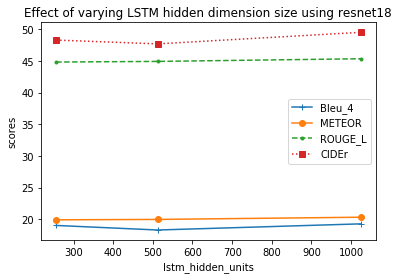}
    \caption{ResNet18}
    \label{subfig:lstm-units-resnet18}
  \end{subfigure}
  \begin{subfigure}[b]{0.45\linewidth}
    \includegraphics[width=\linewidth]{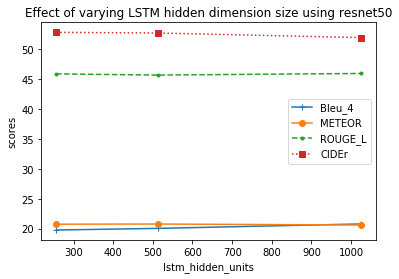}
    \caption{ResNet50}
    \label{subfig:lstm-units-resnet50}
  \end{subfigure}
  \begin{subfigure}[b]{0.45\linewidth}
    \includegraphics[width=\linewidth]{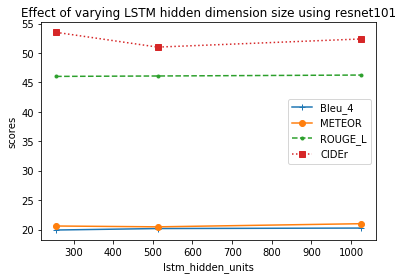}
    \caption{ResNet101}
    \label{subfig:lstm-units-resnet101}
  \end{subfigure}
  \caption{Effect of varying number of LSTM hidden units keeping CNN encoder fixed.}
  \label{fig:exp-lstm-units}
\end{figure}

Lastly, we looked at the effect of varying number of LSTM hidden units by experimenting with 256, 512, and 1024 units. The experiment was carried out by keeping the encoder the same and varying the LSTM units. This experiment was repeated for all encoders: ResNet18, ResNet50, and ResNet101. It was hypothesized that increasing the hidden units would increase captioning quality as the capacity to store information increases and hence, it would be able to model long term dependencies better. The results of this experiment is summarized in Table \ref{tab:lstm-varying-lstm-dimensions}. From Figure \ref{fig:exp-lstm-units}, we can see that the performance improves by a negligible amount as the number of units increase. We suspect this is due to over-fitting as we are using Flickr8k which is a small dataset compared to MSCOCO and Flickr30k.

\begin{table}[htpb!]
  \caption{Experimental results for varying number of hidden units in an LSTM for the CNN+LSTM architecture.}
  \label{tab:lstm-varying-lstm-dimensions}
  \centering
  \begin{tabular}{lllllllll}
    \toprule
    Encoder & LSTM Size & BLEU-1 & BLEU-2 & BLEU-3 & BLEU-4 & METEOR & ROUGE\_L & CIDEr \\
    \midrule
    ResNet18 & 256 & 58.90 & 41.04 & 28.07 & 19.04 & 19.93 & 44.84 & 48.31 \\
& 512 & 58.57 & 40.37 & 27.39 & 18.32 & 19.99 & 44.94 & 47.71 \\
& 1024 & 59.75 & 41.46 & 28.37 & 19.29 & 20.34 & 45.36 & 49.53 \\
\midrule
    ResNet50 & 256 & 60.01 & 42.33 & 29.08 & 19.79 & 20.76 & 45.83 & 52.76 \\
& 512 & 59.95 & 42.34 & 29.26 & 20.06 & 20.79 & 45.63 & 52.65 \\
& 1024 & 60.68 & 42.98 & 29.96 & 20.80 & 20.62 & 45.90 & 51.88 \\
\midrule
ResNet101 & 256 & 60.51 & 42.42 & 29.28 & 19.92 & 20.60 & 46.03 & 53.52 \\
& 512 & 60.73 & 43.06 & 29.55 & 20.17 & 20.46 & 46.10 & 51.01 \\
& 1024 & 60.73 & 42.77 & 29.63 & 20.25 & 20.99 & 46.26 & 52.39 \\
    \bottomrule
  \end{tabular}
\end{table}

\subsection{ResNet + Transformer}
Our baseline model for the ResNet-Transformer architecture was a ResNet18 model with 3 transformer layers. Each layer uses just a single head in the baseline. We conducted three experiments: varying the type of encoder model, varying the number of decoder layers, and varying the number of heads. Additionally we analyzed the effect of finetuning the encoder and seeing the performance of the model after finetuning. All experiments were trained using the Adam optimizer with a fixed learning rate of 0.00004 on a Geforece GTX 1080 Ti GPU. Termination of training was determined by early stopping to obtain the best possible BLEU-4 scores. The results are compiled in Tables 4-7.

\begin{figure}[!htpb]
  \centering
  \includegraphics[width=0.5\linewidth]{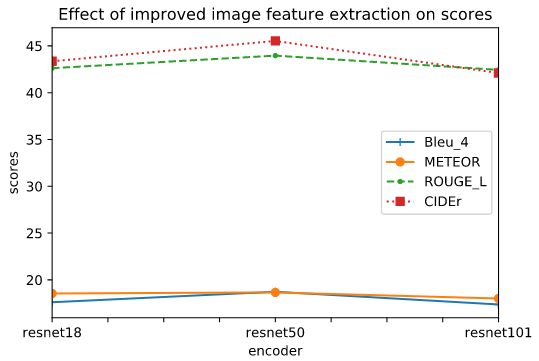}
  \caption{Effect of varying encoders on image captioning for CNN+Transformer architecture.}
  \label{fig:exp-trans-image-quality}
\end{figure}

For analyzing the different encoder models, we used ResNet18, ResNet50 and ResNet101. It was hypothesized that using a larger encoder model should improve the caption generation as the image representation would be better. Surprisngly, from Figure \ref{fig:exp-trans-image-quality}, we see that ResNet50 model performs best whereas the ResNet101 model gives the worst performance. This could be because the dataset size is very small. ResNet18 could be underfitting, whereas ResNet101 could be overfitting. The results are summarized in Table \ref{tab:trans-varying-encoder}.

\begin{table}[!htpb]
  \caption{Experimental results for varying encoder in CNN+Transformer architecture.}
  \label{tab:trans-varying-encoder}
  \centering
  \begin{tabular}{llllllll}
    \toprule
    Encoder & BLEU-1 & BLEU-2 & BLEU-3 & BLEU-4 & METEOR & ROUGE\_L & CIDEr \\
    \midrule
    ResNet18 & 59.08 & 40.35 & 26.87 & 17.60 & 18.55 & 42.62 & 43.36 \\
    ResNet50 & 60.15 & 41.61 & 28.02 & 18.73 & 18.66 & 43.96 & 45.53 \\
    ResNet101 & 59.46 & 40.89 & 26.90 & 17.37 & 18.01 & 42.44 & 42.10 \\
    \bottomrule
  \end{tabular}
\end{table}

\begin{figure}[htpb!]
  \centering
  \begin{subfigure}[b]{0.45\linewidth}
    \includegraphics[width=\linewidth]{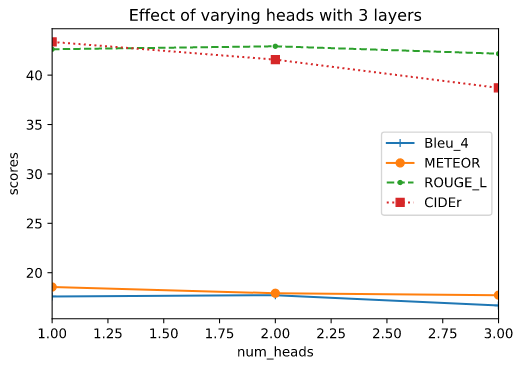}
    \caption{3 layers}
    \label{subfig:3_layers}
  \end{subfigure}
  \begin{subfigure}[b]{0.45\linewidth}
    \includegraphics[width=\linewidth]{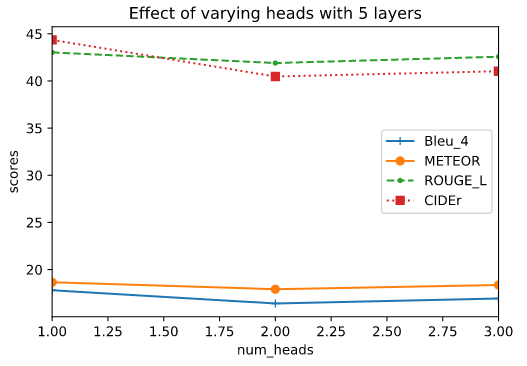}
    \caption{5 layers}
    \label{subfig:5_layers}
  \end{subfigure}
  \begin{subfigure}[b]{0.45\linewidth}
    \includegraphics[width=\linewidth]{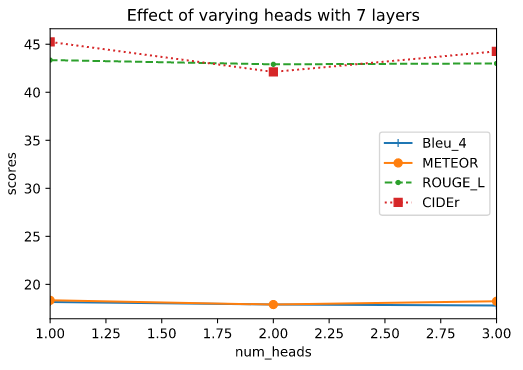}
    \caption{7 layers}
    \label{subfig:7_layers}
  \end{subfigure}
  \caption{Effect of varying number of heads keeping number of decoder layers fixed.}
  \label{fig:varying_heads}
\end{figure}

\begin{figure}[htpb!]
  \centering
  \begin{subfigure}[b]{0.45\linewidth}
    \includegraphics[width=\linewidth]{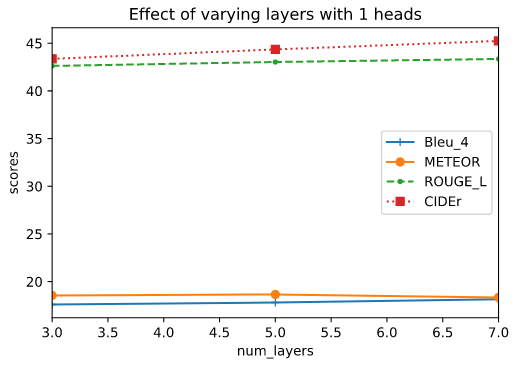}
    \caption{1 head}
    \label{subfig:1_head}
  \end{subfigure}
  \begin{subfigure}[b]{0.45\linewidth}
    \includegraphics[width=\linewidth]{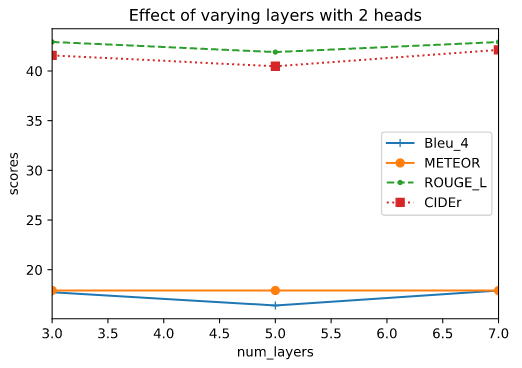}
    \caption{2 heads}
    \label{subfig:2_head}
  \end{subfigure}
  \begin{subfigure}[b]{0.45\linewidth}
    \includegraphics[width=\linewidth]{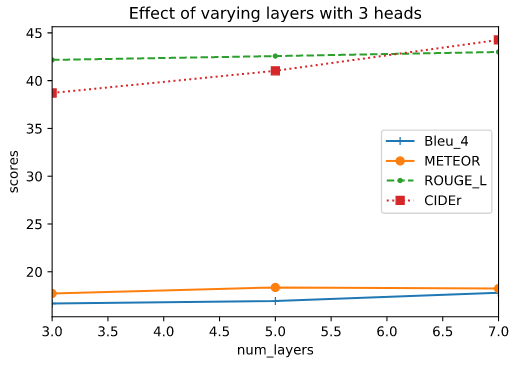}
    \caption{3 heads}
    \label{subfig:3_head}
  \end{subfigure}
  \caption{Effect of varying number of decoder layers keeping number of heads fixed.}
  \label{fig:varying_layers}
\end{figure}

For analyzing the effect of number of decoder layers and heads, we kept the type of encoder fixed as ResNet18. It was hypothesized that increasing the number of heads should improve the quality as the transformer gets to attend to information from different representation subspaces. The results are compiled in Table \ref{tab:trans-varying-heads}. From Figure \ref{fig:varying_heads}, we see that increasing the number of heads had either no effect or was even reducing the performance of the model. It is possible that our learning rate is not tuned perfectly, but we suspect that increasing the number of heads causes the model to overfit as our dataset size is very small compared to bigger datasets like MSCOCO or Flickr30k. Increasing dropout or other regularization techniques might help curb this effect but we have not experimented with this. Similar to changing the number of heads, it was hypothesized that increasing the number of layers should improve performance of caption generation as the transformer learns a better representation of the word as it gets deeper. From Figure \ref{fig:varying_layers}, we see that changing the number of layers also showed no significant or noticeable changes. At times the model performed better and at times worse. These results are compiled in Table \ref{tab:trans-varying-layers}.

\begin{table}[htpb!]
  \caption{Experimental results for varying number of heads keeping decoder layers fixed for the CNN+Transformer architecture. Encoder used is ResNet18.}
  \label{tab:trans-varying-heads}
  \centering
  \begin{tabular}{lllllllll}
    \toprule
    Layers & Heads & BLEU-1 & BLEU-2 & BLEU-3 & BLEU-4 & METEOR & ROUGE\_L & CIDEr \\
    \midrule
    3 & 1 & 59.08 & 40.35 & 26.87 & 17.60 & 18.55 & 42.62 & 43.36 \\
      & 2 & 60.10 & 41.31 & 27.22 & 17.73 & 17.92 & 42.92 & 41.57 \\
      & 3 & 58.86 & 40.24 & 26.05 & 16.68 & 17.73 & 42.17 & 38.71 \\
\midrule
    5 & 1 & 58.87 & 40.38 & 27.06 & 17.81 & 18.65 & 43.03 & 44.35	 \\
      & 2 & 57.25 & 39.04 & 25.62 & 16.40 & 17.91 & 41.90 & 40.47 \\
      & 3 & 57.89 & 39.40 & 26.08 & 16.93 & 18.36 & 42.57 & 41.03 \\
\midrule
    7 & 1 & 60.10 & 41.45 & 27.48 & 18.16 & 18.34 & 43.35 & 45.24 \\
      & 2 & 59.68 & 41.14 & 27.48 & 17.91 & 17.90 & 42.91 & 42.12 \\
      & 3 & 59.84 & 41.18 & 27.27 & 17.80 & 18.24 & 43.00 & 44.26 \\
    \bottomrule
  \end{tabular}
\end{table}

\begin{table}[htpb!]
  \caption{Experimental results for varying number of decoder layers keeping heads fixed for the CNN+Transformer architecture. Encoder used is ResNet18.}
  \label{tab:trans-varying-layers}
  \centering
  \begin{tabular}{lllllllll}
    \toprule
    Heads & Layers & BLEU-1 & BLEU-2 & BLEU-3 & BLEU-4 & METEOR & ROUGE\_L & CIDEr \\
    \midrule
    1 & 3 & 59.08 & 40.35 & 26.87 & 17.60 & 18.55 & 42.62 & 43.36 \\
      & 5 & 58.87 & 40.38 & 27.06 & 17.81 & 18.65 & 43.03 & 44.35 \\
      & 7 & 60.10 & 41.45 & 27.48 & 18.16 & 18.34 & 43.35 & 45.24 \\
\midrule
    2 & 3 & 60.10 & 41.31 & 27.22 & 17.73 & 17.92 & 42.92 & 41.57	 \\
      & 5 & 57.25 & 39.04 & 25.62 & 16.40 & 17.91 & 41.90 & 40.47 \\
      & 7 & 59.68 & 41.14 & 27.48 & 17.91 & 17.90 & 42.91 & 42.12 \\
\midrule
    3 & 3 & 58.86 & 40.24 & 26.05 & 16.68 & 17.73 & 42.17 & 38.71 \\
      & 5 & 57.89 & 39.40 & 26.08 & 16.93 & 18.36 & 42.57 & 41.03 \\
      & 7 & 59.84 & 41.18 & 27.27 & 17.80 & 18.24 & 43.00 & 44.26 \\
    \bottomrule
  \end{tabular}
\end{table}

\begin{figure}[htpb!]
  \centering
  \includegraphics[width=0.5\linewidth]{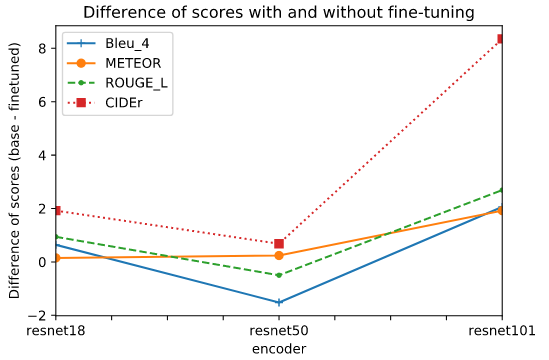}
  \caption{Effect of fine-tuning on image captioning for CNN+Transformer architecture. Positive difference indicates that fine-tuned model outperforms the base model without fine-tuning.}
  \label{fig:exp-trans-finetuned}
\end{figure}

Finally, the most notable and obvious difference is when fine-tuning. It was hypothesized that fine-tuning the encoder would improve the quality of the caption. The summary of the results is listed in Table \ref{tab:trans-finetuned}, and from Figure \ref{fig:exp-trans-finetuned}, it is clear that the fine-tuned encoder models consistently perform better than the encoder models which are not. This makes sense considering that the encoder models are trained on ImageNet whereas our dataset is Flickr8k.

\begin{table}[htpb!]
  \caption{Experimental results for fine-tuning last CNN layer in CNN+Transformer architecture. It shows the difference of scores by subtracting the base model scores from fine-tuned scores.}
  \label{tab:trans-finetuned}
  \centering
  \begin{tabular}{llllllll}
    \toprule
    Encoder & $\Delta$ BLEU-1 & $\Delta$ BLEU-2 & $\Delta$ BLEU-3 & $\Delta$ BLEU-4 & $\Delta$ METEOR & $\Delta$ ROUGE\_L & $\Delta$ CIDEr \\
    \midrule
    ResNet18 & 1.18 & 1.36 & 0.94 & 0.64 & 0.15 & 0.94 & 1.92 \\
    ResNet50 & 0.20	& -0.34	& -1.05	& -1.52	& 0.24 & -0.50 & 0.68  \\
    ResNet101 & 2.09 & 2.57	& 2.69 & 2.06 & 1.92 & 2.69 & 8.35 \\
    \bottomrule
  \end{tabular}
\end{table}

\section{Conclusions}

We have presented and compared two different architectures for image caption generation. A ResNet model is used as an image feature extractor. For decoding we experimented with LSTMs and Transformers. We performed a sensitivity analysis of various hyperparameters. Experiments show that fine-tuning the encoder model almost always improves the outcome of the decoder model. The LSTM model using ResNet50 and ResNet101 surpass our reference baseline, the Soft-Attention Model even without finetuning. This could be due to the higher representative capability of the ResNet model as compared to the VGG model used in the Soft-attention model. All the other models surpass the baseline upon finetuning, with the only exception being using Resnet50 and a Transformer decoder. We have shown the effect of tuning other hyperparameters such as the number of hidden units for LSTMs, the number of decoder layers and number of heads used in multi-head attention for Transformers. It was noticed that increasing the number of heads, layers or hidden vector size does not always improve results and may even result in reducing the output quality. We believe that this is due to model overfitting as our dataset size is very small. A natural continuation of this work would be to experiment with larger datasets such as Flickr30k or MSCOCO. Apart from that, it would also be interesting to experiment with changing the word embedding size used in the LSTM model, or even finding out the effect of using pre-trained word embeddings such as GloVe vectors. Finally, we could experiment with more complicated Transformer architectures such as those by \citet{transformer-stacked-attention}.

\subsubsection*{Acknowledgments}
The authors would like to thank Jimmy Ba for his thoughtful instruction throughout the Neural Networks course as well as providing helpful insights during our project consultation meetings.

% \subsubsection*{Contributions}
% The contributions from both authors was equal in the project. 

% In terms of implementation, Aravind Varier coded up the base models used for training while Amish Patel coded up the steps involved in testing/validation in parallel and the data visualization for the results gathered. The code for this project is available \href{https://github.com/aravindvarier/Image-Captioning-Pytorch}{\textbf{here}}.

% All the CNN+LSTM experiments were carried out by Amish Patel and the CNN+Transformer experiments were carried out by Aravind Varier.

% The writing portion was distributed equally as well. Amish Patel was responsible for the Introduction and the sections related to CNN+LSTM experiments while Aravind Varier was responsible for sections related to CNN+Transformer and Conclusion.

\small
\bibliographystyle{plainnat}
\bibliography{references}

\medskip
\end{document}